\documentclass{article}

\usepackage[preprint,nonatbib]{neurips_2018}

\usepackage{graphicx}
\usepackage{enumitem}   

\usepackage{caption}
\usepackage{subcaption}

\usepackage[utf8]{inputenc} % allow utf-8 input
\usepackage[T1]{fontenc}    % use 8-bit T1 fonts
\usepackage{hyperref}       % hyperlinks
\usepackage{url}            % simple URL typesetting
\usepackage{booktabs}       % professional-quality tables
\usepackage{amsfonts}       % blackboard math symbols
\usepackage{amsmath}
\usepackage{array, makecell}

\usepackage{amssymb}
\usepackage{nicefrac}       % compact symbols for 1/2, etc.
\usepackage{microtype}      % microtypography
\usepackage{lipsum}
\usepackage{multicol}
\usepackage[linesnumbered,ruled,vlined]{algorithm2e}

\newcolumntype{x}[1]{>{\let\newline\\\arraybackslash\hspace{0pt}}p{#1}}

\title{Correlation Based Feature Subset Selection for Multivariate Time-Series Data
}

\author{
   Bahavathy Kathirgamanathan \\
    School of Computer Science\\
   University College Dublin\\
   \texttt{bahavathy.kathirgamanathan@ucdconnect.ie}\\
   \And
     P\'{a}draig Cunningham \\
  School of Computer Science\\
  University College Dublin\\
  \texttt{padraig.cunningham@ucd.ie} \\
 }
 
\begin{document}
\maketitle

\begin{abstract}
Correlations in streams of multivariate time series data means that typically, only a small subset of the features are required for a given data mining task. In this paper, we propose a technique which we call Merit Score for Time-Series data (MSTS) that does feature subset selection based on the correlation patterns of single feature classifier outputs. We assign a Merit Score to the feature subsets which is used as the basis for selecting `good' feature subsets. The proposed technique is evaluated on datasets from the UEA multivariate time series archive and is compared against a Wrapper approach for feature subset selection. MSTS is shown to be effective for feature subset selection and is in particular effective as a data reduction technique. MSTS is shown here to be computationally more efficient than the Wrapper strategy in selecting a suitable feature subset, being more than 100 times faster for some larger datasets while also maintaining a good classification accuracy.
%\keywords{Feature Subset Selection \and Time Series Classification \and Merit Score \and Multivariate Time Series}
\end{abstract}

\section{Introduction}
\label{intro}
Multivariate time series (MTS) data is common in many real-world applications such as Human Activity Recognition \cite{Seto2015} and for clinical diagnosis based on EEG or ECG signals. Many researchers have recently focused on univariate time series, however in many real world applications it may be more common to encounter MTS problems. The key added complexity of MTS data is in the interaction between the streams of data \cite{Pasos2020}. Due to the time series nature of the data, streams of data in MTS problems are often correlated which can lead to redundant data. Feature subset selection (FSS) is used to pre-process data and identify a subset of the original features to take forward into the data mining task. Hence, FSS aims to remove any redundant data. FSS is done with three key objectives \cite{IsabelleGuyon2003}:
\begin{itemize}
\item[--] Improve classification performance by including features with high predictive capabilities and excluding the redundant features.
\item[--] Reduce computational expense by reducing the feature subset size
\item[--] Provide insights into the data through the features selected
\end{itemize}
In FSS of MTS data, a feature refers to an univariate time series \cite{Ircio2020,Wang2013,Yanga,Dhariyal2021}. Many current approaches use feature extraction where the time series is transformed or derived and these extracted values are termed as the features. However, this can lead to a loss of information from the original time series during the extraction process.  Furthermore, FSS allows the user to reduce data collection costs by excluding any discarded features from subsequent data collection and processing. In feature extraction, all the original features will still need to be collected and little insight into the more important features for the data mining task can be obtained. An example of this, is in activity recognition where wearable sensors are placed on the body to capture motion. FSS on this data can reduce costs associated with data collection, as the sensors which provide redundant data may not need to be mounted.
\par
As MTS data generally contains correlations between the features, correlation based approaches may work well to remove the redundant features. Correlation based Feature Selection (CFS) is a technique that has been used successfully outside of the time series domain for selecting a feature subset from multivariate data \cite{hall1999correlation}. It is not however directly applicable to time series as it requires the data in a feature vector representation. In this paper, we propose a feature subset selection method for working with multivariate time series where we use outputs of single feature classifiers to calculate correlations in the data. The methods presented here builds upon previous work where we proposed the correlation based feature selection technique along with some preliminary evaluations \cite{kathirgamanathan2020feature}.
\par
In this paper, a correlation based feature subset selection technique for multivariate time series data is proposed where single feature classifier outputs are used to evaluate correlations.%\footnote{All code used for the evaluations will be provided with this paper as a GitHub repository}. 
In Section \ref{sec:FS} an overview of feature selection  techniques is presented with a focus on techniques used in the time series space. Section \ref{sec:TTSC} gives an overview of the State-of-the-art (SOTA) classification techniques for times series that will be used for evaluating the feature subsets. Section \ref{sec:CB_FS} gives an overview of correlation based feature selection and introduces our proposed methods. Section \ref{sec:ES} goes on to present the evaluation strategy undertaken, Section \ref{Sec:Res} shows the results from the evaluations and finally the conclusions are presented in Section \ref{sec:conc}. 
\section{Feature Selection}
\label{sec:FS}
Given a dataset with $n$ dimensions, there will be $2^n$ feature subsets that can be generated. Feature Selection techniques aim to identify good feature subsets in this space. Generally, feature selection techniques are classified as Filter methods, Wrapper methods, or Embedded methods \cite{IsabelleGuyon2003,CarlosMolina2002}. Filters choose feature subsets independent of the classifier and is usually done as a pre-processing step. Wrappers use the learning algorithm by using predictive accuracy as a score for selecting feature subsets. Embedded methods include feature selection in the training process and are usually specific to certain learning algorithms. Filter methods, while computationally not expensive,  are less accurate. Wrapper methods on the other hand are more effective as they evaluate the performance in context, however they are computationally expensive. Although the best strategy usually depends on the task at hand, Wrappers are often used as a preferred feature selection method \cite{Cunningham2021}. Although feature selection is well established outside the time series domain, many of these techniques are not directly applicable to time series data due to the extra dimension that is present in time series data. Hence, there has been less progress made in feature selection for time series.
\subsection{Feature Subset Selection for Time Series}
\label{sec:FSSTS}
\par
A time series is a sequence of observations that is time based, $x_i(t); [i=1,\ldots, n; t=1,\ldots,m]$, where $i$ indexes the data gathered at time point $t$. The time series is univariate when $n$ is 1 and multivariate when $n$ is greater than or equal to 2. When working with time series data, the time series could be of different lengths and in this case $m$ may not be the same for all samples. Multivariate time series can often be large in size and hence it is important to have suitable methods for preprocessing the data prior to classification. 
\par
There are two commonly used techniques to deal with the high dimensionality of MTS. These are feature extraction and feature subset selection. Feature extraction methods involve the transformation or mapping of the original data into extracted features. Feature subset selection involves reducing the number of features from the original dataset that is used for analysis by selecting only the features required and removing the redundant features. A drawback with using feature extraction methods is that there can be a loss of information compared to feature subset selection which uses the original time series. Here the focus will be on feature subset selection methods as our work focuses on this rather than feature extraction.
\par
Many state of the art feature subset selection techniques such as Recursive Feature Elimination (RFE) require each item to be inputted in the form of a column vector \cite{IsabelleGuyon2003}. Multivariate time series are naturally  represented as a $m \times n$ matrix which makes these methods not ideal when working with multivariate time series. This is in particular the case when considering correlation based feature selection as vectorising time series data will lead to a loss of correlation related information between the features. Hence, although there has been a lot of work undertaken outside of the time series domain for multi-variate feature selection, there is limited work in feature selection for multivariate time series (MTS).
\par
Some correlation based methods have been implemented for feature subset selection in time series. Many methods typically used to calculate correlation such as Spearman's correlation can be effective for non-time series data however has been shown to produce poor results when implemented on time series \cite{Wang2013}. In one study, Pearson's correlation and the Symmetrical uncertainty scores are used together to capture linear and non-linear relationships between the features and classes \cite{Saikhu2019}. Here,  the authors use these relationships to calculate a Merit Score which becomes the basis for selecting feature subsets. However as the symmetrical uncertainty only works where the data is discrete, the authors discretise the time series prior to calculating correlation which can lead to the loss of information and hence may not ideal for working with time series data.
\par
Principal Component Analysis (PCA) is a dimension reduction technique that has been used as part of multivariate feature selection. PCA allows correlation information between variables to be preserved. CLeVer is a technique which utilises properties of the descriptive common principal components for MTS feature subset selection. This method uses loadings to weight the contribution of each feature to the principal components. By ranking each feature by how much it contributes to the principal components, this method aims to reduce the dimensionality while retaining information related to both the original features and the correlation amongst the features \cite{Yanga}. A further recent development for channel selection uses the class-wise separation based on centroid-pairs where each class centroid is used to represent that class and the distance between the centroids is used to determine the separability \cite{Dhariyal2021}. This study focuses on the scalability and shows that reducing the feature size can reduce computational cost required when using popular classifiers for multivariate time series classification tasks.
\subsection{Mutual Information based methods}
Mutual Information (MI) is another well-known technique that has been used on MTS data to measure correlation. MI is advantageous over other methods as it allows for both linear and nonlinear correlation to be captured. MI is a concept that is used in information theory and it is based on a measure of the uncertainty of random variables, termed Shannon's entropy \cite{Shannon1948}. MI has been used successfully for feature selection \cite{Doquire2012}. As the MI between two random variables increases, it is expected the correlation between them will also be greater.
\par
Formally defining the concepts behind MI, given two continuous random variables $X$ and $Y$, the entropy of X is defined as:
\newline
\begin{equation}
    H(X) = - \int p(x)\:log\:p(x)\:dx
\end{equation}
The entropy of $X$ and $Y$ is defined as:
\newline
\begin{equation}
    H(X,Y) = - \iint p(x,y)\:log\:p(x,y)\:dx\:dy
\end{equation}
The MI between $X$ and $Y$ is defined as:
\newline
\begin{equation}
    MI(X;Y) = \iint p(x,y) \: log\:\frac{p(x,y)}{p(x)p(y)}\:dx\:dy
\end{equation}
where $p(x,y)$ is the joint probability density function of X and Y and $p(x)$ and $p(y)$ are the probability density function of X and Y respectively. Hence MI and entropy can be combined in the form:
\begin{equation}
    MI(X;Y) = H(X)\:+\:H(Y)\:-H(X,Y)
\end{equation}
%To deal with the dimensionality of time series techniques such as MI based on K-nearest neighbours and MI with class separability have been used. 
There are few studies that have been done using MI for feature selection of time series. The class separability based feature selection (CSFS) algorithm uses MI between the original variables as features for classification. Based on this, the ratio of between class scattering to within class scattering is used to identify the contribution of a feature to the classification, hence allowing the original variables to be ranked according to their contribution to the classification \cite{Han2013}. 
\par
MI is typically calculated in a pairwise manner which may not be ideal when working with multidimensional data. To avoid this, some studies have used a k-nearest neighbour ($k$-NN) approach to calculate the MI which avoids the need to calculate the probability distribution function and therefore can be used on the original multidimensional feature subset \cite{Han2015,Liu2016}. In a very recent development \cite{Ircio2020}, the authors propose a Filter method for feature subset selection where they assign a score function to assess the relevance of each feature subset. Mutual information based on a $k$-NN strategy is used to measure the information shared between the two time series. To measure the MI between the class and the time series, methods which do not assume any specific probability function are used as the computation of MI between a class and time series is still an unsolved problem. Furthermore, this approach has been evaluated on a subset of datasets from the UEA repository and has shown potential for feature subset selection techniques based on MI. This was the only evaluation found in literature, to the best of our knowledge, which presents a feature subset selection technique which selects a subset of the original time series and presents an evaluation on multiple datasets suggesting there is scope for work to be conducted in time series based feature subset selection.
\section{Time Series classification}\label{sec:TTSC}
Classification techniques used in time series can be grouped into categories such as distance based, shapelet based, dictionary based, deep learning methods, ensembles, and a recently developed technique called ROCKET which does not strictly belong in any of the other categories. Recent literature has seen the development of thorough literature surveys evaluating these techniques for time series classification \cite{Bagnall2017,Pasos2020,IsmailFawaz2019}. Based upon these reviews, we focus on the 1NN-DTW and ROCKET classifiers. 1NN-DTW is known to be a reliable benchmark whereas ROCKET is currently the state-of-the-art classifier for multivariate time series classification. 
\subsection{1NN-DTW}
\par
1-Nearest Neighbour (1-NN) with Dynamic Time Warping (DTW) is known as one of the most reliable and simple approaches for time series classification and is frequently used as a benchmark classifier \cite{Bagnall2017}. Despite this, 1NN-DTW has a high run-time making it unsuitable for many applications. 
\par
When using DTW for multivariate problems, there are a few different strategies that have been employed \cite{Shokoohi-Yekta2017}. The independent warping ($DTW_I$) strategy treats each dimension separately by calculating a distance matrix, M for each dimension and then uses the sum of the DTW distances for classification. Dependent warping ($DTW_D$) takes the same warping across all dimensions by calculating the Euclidean distance between the two vectors representing all the series rather than the single series and redefining this as a matrix. The DTW function then uses this redefined matrix. This technique assumes that there is some relation between the multivariate time series. It has been shown that neither performs better than the other but instead depends on the problem at hand \cite{Shokoohi-Yekta2017}.
\subsection{ROCKET + Variants}
ROCKET \cite{Dempster2020} is a recent development which has shown great potential. ROCKET borrows ideas from deep neural networks where a simple linear classifier is trained on random convolutional kernels. ROCKET works by generating random kernels which are convolved along the time series to produce a feature map. Two features are then extracted for each kernel, the global max pooling and the proportion of positive values (PPV). These extracted features then become the transformed data which is used inside the linear classifier. This development has been shown to achieve state-of-the-art performance whilst maintaining a lower computational load than other state-of-the-art techniques.
\par
MiniRocket was later developed as a reformulated version of ROCKET \cite{Dempster2020Mini}. MiniRocket is faster than ROCKET and almost deterministic. It differs from ROCKET where it uses a fixed set of kernels instead of random kernels. MiniRocket only extracts the PPV as this was found to be more important the the max pooling feature. MultiRocket was very recently developed and works by expanding the set of features produced  from the transform in MiniRocket \cite{Tan2021}. MultiRocket improves accuracy while compromising slightly on computational expense. However this technique is still considerably faster than other algorithms of similar accuracy.
\par
The key selling point of ROCKET and its two variants is its low computational expense. Most accurate methods for TSC to date have had a high computational complexity. ROCKET and its variants only require a fraction of this time. In this paper, MiniRocket is used in our evaluation as it is the suggested default variant of ROCKET \cite{Dempster2020Mini}.

\section{Correlation Based Feature Selection}\label{sec:CB_FS}
Correlation Based Feature Selection (CFS) is a technique which ranks feature subsets based on a Merit Score calculated from the correlations in the data \cite{hall1999correlation}. The heuristic used relies on the principle that ``a good feature subset is one that contains features highly correlated to the class, yet uncorrelated with each other". In CFS, a Merit Score is calculated using feature-class and feature-feature correlations which becomes the score by which individual feature subsets are evaluated. The feature-class correlation indicates the level in which the feature represents the class while the feature-feature correlation indicates any redundancies there may be between the features. The Merit Score $M_S$ is defined as follows:
\begin{equation}
    M_S = \frac{k\overline{r_{cf}}}
    {\sqrt{k+k(k-1)\overline{r_{ff}}}}
    \label{eqn:MS}
\end{equation}
Where $\overline{r_{cf}}$ is the average correlation between the features in the subset and the class label and $\overline{r_{ff}}$ is the average correlation between the selected features. $k$ represents the number of features in the subset.
\par
To measure the correlations, techniques such as symmetrical uncertainty based on information gain, feature weighting based on the Gini-index, or the Minimum Description Length (MDL) principle are used \cite{hall1999correlation}. Information gain based methods have been shown to work well and hence has been used often in CFS implementations. CFS generally works using correlations calculated between the features themselves. This is not feasible while working with time series as it requires the data in a feature vector format. 
\subsection{Proposed Method}
In our proposed technique where feature selection is carried out based on a Merit Score for time series data (MSTS), we use single feature predictions of the class labels to calculate the correlation scores. To calculate the merit of feature subsets for time series data, a modified version of Equation \ref{eqn:MS} is defined as follows:
%\begin{figure}
%  \includegraphics[keepaspectratio, width=8cm]{Images/Mushroom_Corr_final.png}
%\caption{Merit Scores calculated using the original technique based on symmetrical uncertainty vs that from single feature predictors for the mushroom dataset}
%\label{fig:HeatMap_Corr}
%
%\end{figure}
\par
\begin{equation}
    MSTS = \frac{k\overline{Y_{cf}}}
    {\sqrt{k+k(k-1)\overline{Y_{ff}}}}
    \label{eqn:MSTS}
\end{equation}
where $Y_{cf}$ and $Y_{ff}$ are correlations calculated on the class labels predicted for the training data rather than on feature values as is the case in Equation \ref{eqn:MS}. Hence, $Y_{cf}$ was calculated by averaging the feature-class correlations of all the features present in the subset. In this paper, the adjusted mutual information (AMI) score is used to calculate the correlations. The AMI score is an adjustment of the MI score to account for chance \cite{Vinh2010}. The AMI score is defined as:
\begin{equation}
    AMI(X,Y) = \frac{MI(X,Y) - E[MI(X,Y)]}{mean(H(X),H(Y))-E[MI(X,Y)]}
    \label{eqn:AMI}
\end{equation}
The `best' feature subset would ideally be the one which has the largest Merit Score from Equation \ref{eqn:MSTS}. Following the Merit Score calculation, a search strategy of the users choice can be used to select this `best' feature subset. Here the evaluation would be very similar to that of a traditional Wrapper strategy but rather than evaluating based on classification performance, the feature subsets will be evaluated based on their assigned merit scores. The search algorithms used for evaluation in this paper are described in more detail in Section \ref{sec:MSSS}.

\subsubsection{Proposed Method on tabular Data}
\begin{figure}[h]
\begin{subfigure}{.5\textwidth}
  \centering
  % include first image
  \includegraphics[width=1\linewidth]{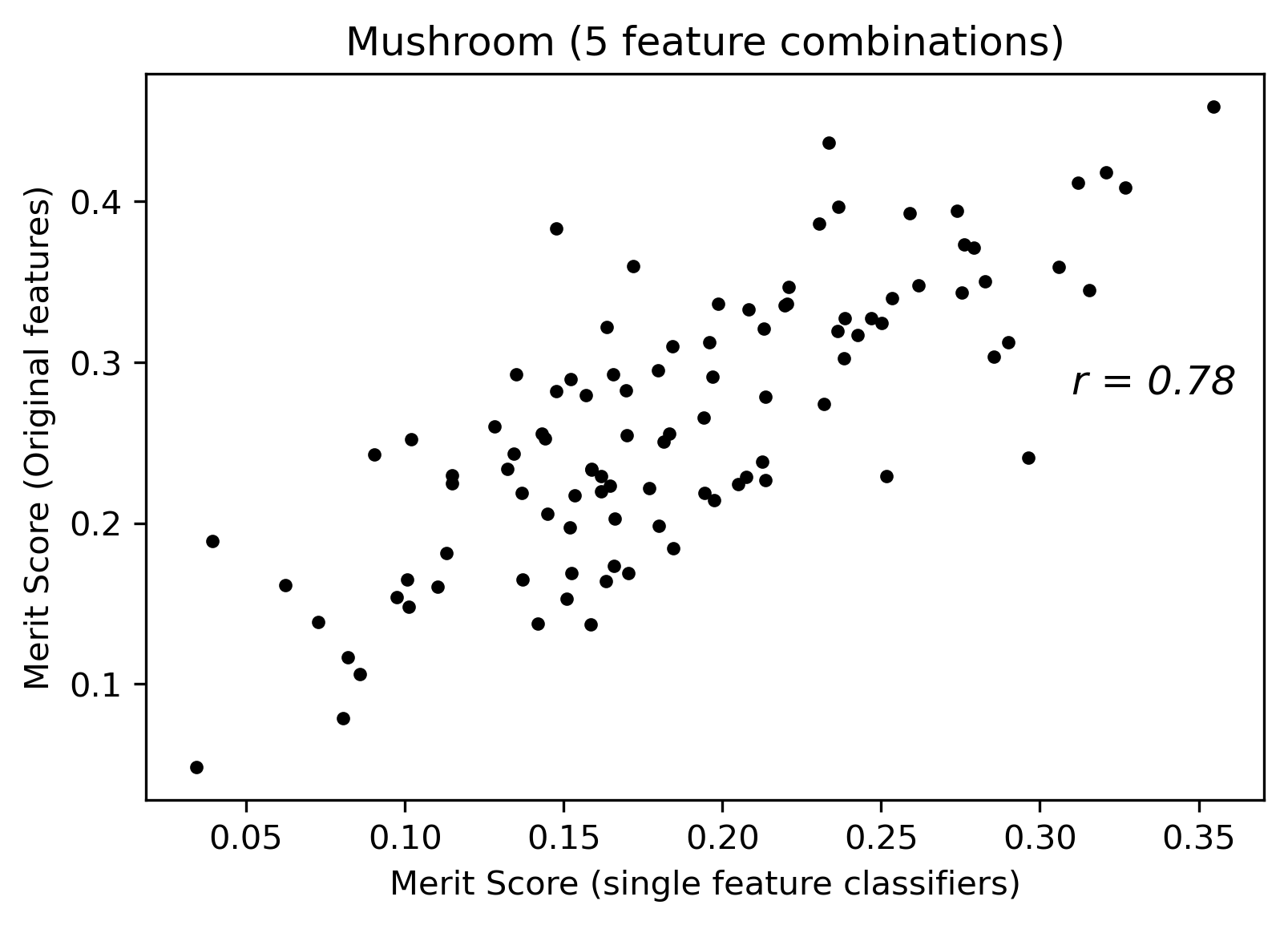}  
  \caption{}
  %\label{fig:sub-first}
\end{subfigure}
\begin{subfigure}{.5\textwidth}
  \centering
  % include second image
  \includegraphics[width=1\linewidth]{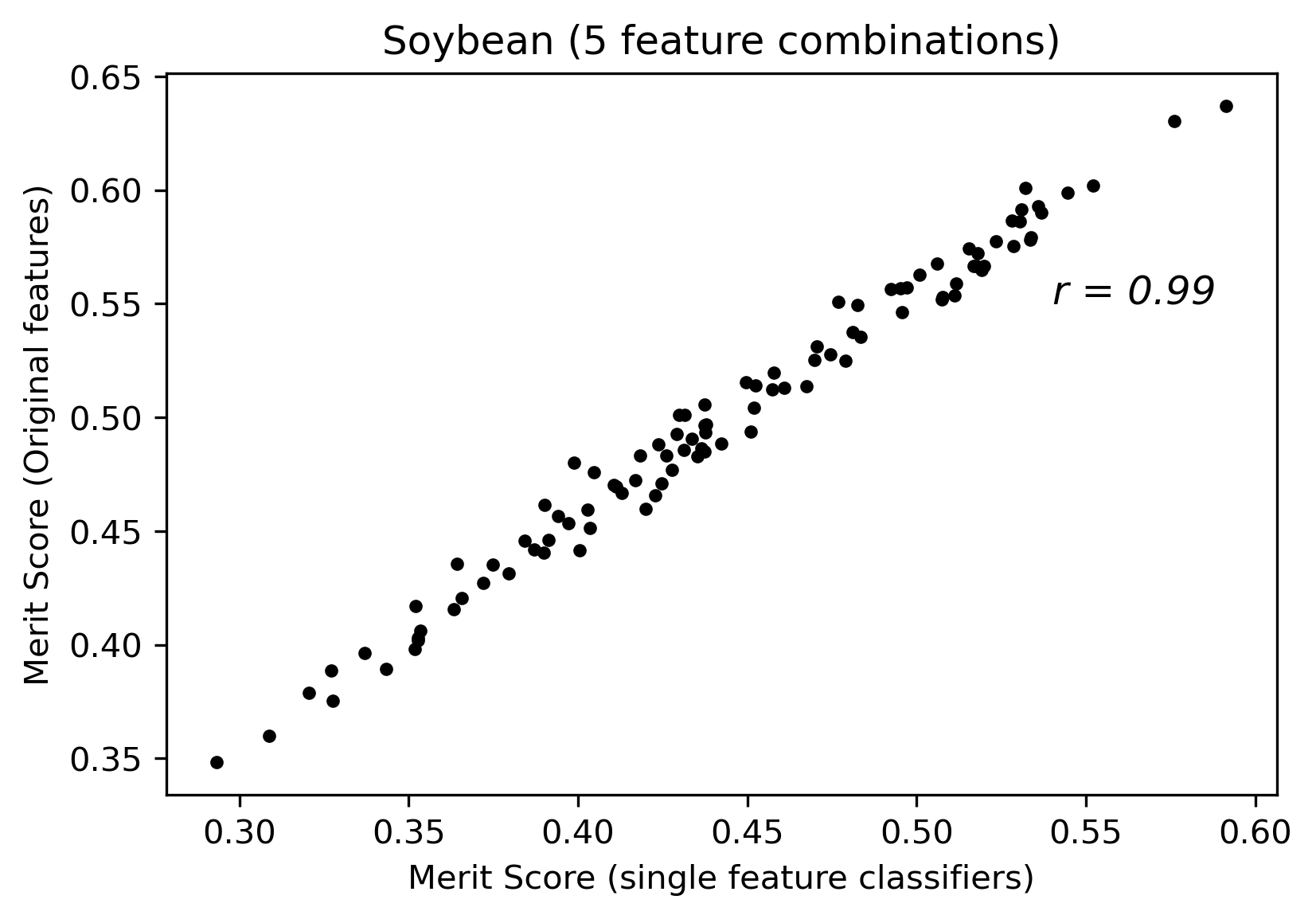}  
  \caption{}
  %\label{fig:sub-second}
\end{subfigure}
\caption{Merit Scores calculated using the original technique vs our proposed technique on 100 randomly selected 5 feature subsets for a)Mushroom dataset, b) Soybean dataset.}
\label{fig:corrcomp}
\end{figure}

To check the effectiveness of using classifier outputs rather than the features themselves for CFS, the proposed technique where we use single feature classifier outputs to calculate correlation scores is tested on two non-time series datasets. The Mushroom and Soybean datasets were both taken from the UCI Machine Learning Repository \cite{Dua:2019} and were datasets that were originally used in \textit{Hall's} work \cite{hall1999correlation}. Figure \ref{fig:corrcomp} shows that the two techniques of calculating Merit Score (original technique versus proposed technique) are highly correlated. For the Merit Score based on the correlations of the original features, symmetrical uncertainty was used to calculate correlation and for our proposed classifier output based method, the AMI score was used. The Soybean dataset in particular shows near perfect correlation. This motivates the use of the classifier output based technique for use in time series as it is not possible to easily calculate correlations between the features themselves.

\subsubsection{Proposed Method on Time Series Data}
\begin{figure}
\centering
  \includegraphics[keepaspectratio, trim = 110 80 120 50, clip,width=0.8\textwidth]{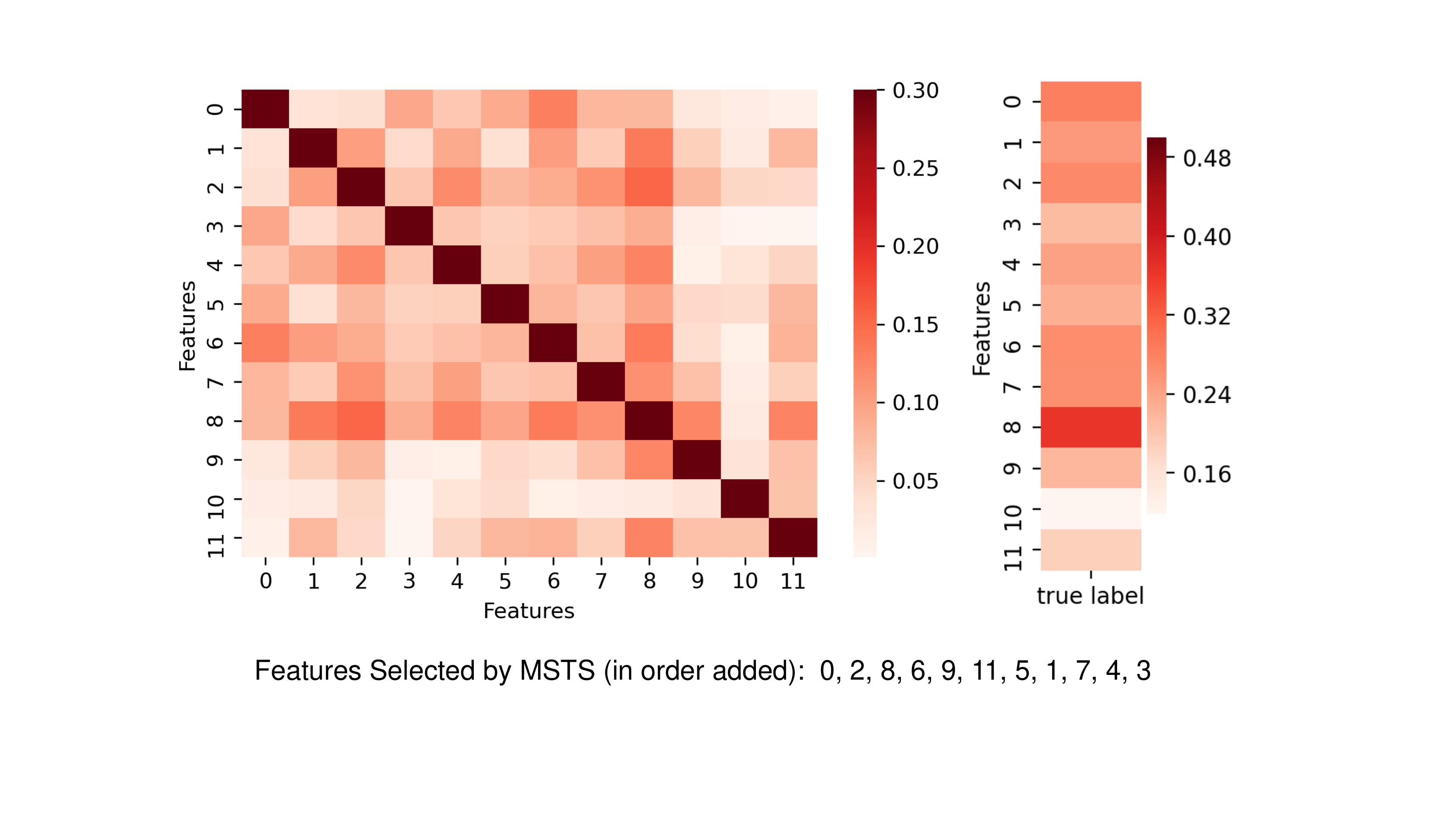}
\caption{Feature-Feature and Feature-Class correlations for the JapaneseVowels Dataset}
\label{fig:HeatMap_Corr}
\end{figure}

To further illustrate the effectiveness of this proposed technique, the feature-feature and feature-class correlations are plotted for time series data. Figure \ref{fig:HeatMap_Corr} shows the feature-feature and feature-class correlations for the JapaneseVowels dataset. In this example, CFS would select feature combinations where the feature-feature correlation matrix (Figure \ref{fig:HeatMap_Corr}a) has a low value and the feature-class correlation vector (Figure \ref{fig:HeatMap_Corr}b) has a high value. In this example, despite feature 8 having the highest feature-class correlation, features 0 and 2 are selected first as there is more merit to using these two features together than using feature 8 with any other feature as both features 0 and 2 have a good predictive capability whilst also being able to bring different information to the classification problem than the other.

\section{Evaluation Strategy}\label{ES}
\subsection{Datasets}\label{sec:datasets}
Our evaluation of the feature selection technique was carried out on datasets from the UEA multivariate Time Series Classification archive \cite{Bagnall2018}. This archive was developed as a collaborative effort between researchers at the University of East Anglia and the University of California. Datasets with less than 4 features were excluded as there is little merit to selecting feature subsets in these datasets. Although, many of the larger datasets were considered to ensure that a variety of datasets were tested, a few large datasets were left out of the analysis due to the high computational costs required to test on all of them. Altogether 19 datasets were selected and a summary of these datasets are provided in Table \ref{tab:datasets}. The datasets come from a variety of domains such as Human Activity Recognition, Motion Classification, and Audio Spectra Classification.
\par
The evaluation was conducted using the default train/test split in the UEA dataset which allows for easier comparison with benchmark results in literature. The feature selection process was carried out on the training data and the performance of the selected feature subsets was evaluated on the test data.
\begin{table}
%\begin{center}
\begin{tabular}{ |x{5.1cm}|x{1.8cm}|x{1.6cm}|x{1.8cm}|x{3.8cm}|}
\hline
Dataset & Dimensions & TS length & \# of Classes\\ 
\hline
ArticularyWordRecognition (AWR) & 9 & 144 & 25\\ 
\hline
FingerMovements (FM) & 28 & 50 & 2\\
\hline
JapaneseVowels (JV) & 12 & 29 & 9\\ 
\hline
NATOPS & 24 & 51 & 6\\ 
\hline
RacketSport (RS) & 6 & 30 & 4\\ 
\hline
SelfRegulationSCP1 (SCP1) & 6 & 896 & 2\\ 
\hline
Heartbeat (HB) & 61 & 405 & 2\\ 
\hline
BasicMotions (BM) & 6 & 100 & 4\\
\hline
HandMovementDirection (HMD) & 10 & 400 & 4\\ 
\hline
Cricket (Cr) & 6 & 1197 & 12\\
\hline
StandWalkJump (SWJ) & 4 & 2500 & 3\\ 
\hline
SelfRegulationSCP2 (SCP2) & 7 & 1152 & 2\\ 
\hline
PEMS-SF & 963 & 144 & 7\\ 
\hline
LSST & 6 & 36 & 14\\ 
\hline
EigenWorms (EW) & 6 & 17984 & 5\\
\hline
ERing (ER) &  4 & 65 & 6\\
\hline
Phoneme (Ph) & 11 & 217 & 39\\
\hline
SpokenArabicDigits (SAD) & 13 & 93 & 10\\
\hline
MotorImagery (MI) & 64 & 3000 & 2\\
\hline
\end{tabular}
\caption{Summary of the datasets used for evaluation}\label{DatasetsTable}
%\end{center}
\label{tab:datasets}
\end{table}
\subsection{Search Strategy}
\label{sec:MSSS}
 The Merit Score gives each feature subset a score that can be used to rank the subsets, however to select the best feature subset, a search strategy must be used as would be used in a Wrapper approach. Evaluating all the feature subsets based on their merit can be computationally expensive. Hence rather than exhaustively searching all feature subsets, a greedy search is used. Sequential Forward Selection (SFS) is a widely used greedy search technique for Wrapper techniques and hence it is used in this evaluation. SFS works by adding a feature one at a time and evaluating the best feature subset at each stage. Hence, for our evaluation initially all 2 feature subsets are evaluated, then the best feature subset is selected and all 3 feature subsets which include the best 2 feature subset are subsequently evaluated. This procedure continues until the Merit Score stops improving upon the addition of a feature. The MSTS strategy used in this evaluation is presented in Algorithm \ref{MSTSalgo}. 
 %The DTW matrix is calculated and stored in advance. DTW allows for a mapping of the time series in a non-linear way and works to find the optimal alignment between both series. DTW can be considered as a one-to-many mapping \cite{Sakoe1978}. As this is a computationally expensive task and will be repeatedly used for cross-validation, it is calculated and stored in advance. 
 \begin{algorithm}[h]
\SetAlgoLined
\SetKwInOut{Input}{Input}
\Input{ $m$ x $n$ feature matrix ($X$), Class labels ($Y_{c}$)}
$Y_f \leftarrow classifier(X, Y_c)$, Make single feature predictions\\
 $Y_{ff} \leftarrow AMI(Y_{f},Y_{f})$, Calculate all feature-feature correlations (Equation \ref{eqn:AMI})\\
 $Y_{cf} \leftarrow AMI(Y_{c}, Y_{f})$, Calculate all class-feature correlations (Equation \ref{eqn:AMI})\\
 Initialise; $n \leftarrow 2$, $\Delta MS \leftarrow 1$\\
 \While{$\Delta MS > 0$}{
    Identify unique n sized feature subsets which includes the best (n-1) sized feature subset\\
    Calculate MeritScore using Equation \ref{eqn:MSTS} for each n sized subset in consideration\\
    $MS_n \leftarrow Max(MeritScore)$\\
    $\Delta MS  \leftarrow MS_n - MS_{n-1}$\\
    $n \leftarrow n+1$\\
 }
 \caption{MSTS: Merit Score based Feature subset selection for MTS}\label{MSTSalgo}
\end{algorithm}
\par
 Class label predictions are first made using individual feature classifiers whereby each individual feature is used one at a time to make class predictions. This uses a 10-fold cross validation and can be done using any classifier of the users choice. In our evaluation, we use 1-NN DTW and MiniRocket. Individual feature-feature and feature-class correlations are then calculated using the AMI score (Equation \ref{eqn:AMI}) based on the class predictions. All unique feature subsets of size \textit{n} where \textit{n} is the initial subset size that is to be test ($n=2$) are then identified and the merit score is calculated for each of these subsets (Equation \ref{eqn:MS}). From the calculated merit score, the n sized feature subset with the highest merit score is selected and then all unique feature subsets of size $n+1$ which include the selected n sized feature subset are identified. The Merit Scores for this new set of feature subsets are then calculated and the evaluation continues until the Merit Score stops increasing when a feature is added as at this point there is no more merit to adding features. The highest Merit Score at the point where the merit stops increasing is selected as the `best' feature subset. This subset is then used on the test data to evaluate the overall performance MSTS. 
\par
In our evaluation, MSTS is compared against a Wrapper strategy which uses a similar greedy forward search but evaluates based on accuracy (WS). The WS strategy used is presented in Algorithm \ref{AlgoGS}. This was used as comparison to the correlation based technique as it is a method commonly used for feature selection in machine learning. The Wrapper strategy was setup identical to the Merit Score approach with the difference being the criteria on which feature subsets are selected where the accuracy from a 10-fold cross validation is used as the scoring criteria. The Wrapper strategy also has one less step as there is no requirement to calculate single feature classifier output.
\begin{algorithm}[h]
\SetAlgoLined
\SetKwInOut{Input}{Input}
\Input{$m$ x $n$ feature matrix ($X$), Class labels ($Y_{c}$)}
$A_f \leftarrow classifier(X, Y_c)$, Calculate single feature accuracy\\
 Initialise; $n \leftarrow 2$, $\Delta WS \leftarrow 1$\\
 \While{$\Delta WS > 0$}{
    Identify unique n sized feature subsets which includes the best (n-1) sized feature subset\\
    Calculate cross-validated accuracy for each n sized subset in consideration\\
    $WS_n \leftarrow Max(Accuracy)$\\
    $\Delta WS  \leftarrow (WS_n - WS_{n-1})$\\
    $n \leftarrow n+1$\\
 }
 \caption{WS: Wrapper strategy }\label{AlgoGS}
\end{algorithm}
\subsection{Evaluation Details}\label{sec:ES}
To evaluate the MSTS feature subset selection method, 19 datasets from the UEA archive as described in Section \ref{sec:datasets} were used. MSTS and a Wrapper strategy were compared against each other and with the benchmark accuracy obtained when using all the features. Both MSTS and the Wrapper feature selection algorithms can be used with any classifier of the users choice. We evaluate and present the results for two separate evaluations using a 1NN-DTW classifier and the MiniRocket classifier.
\par
The evaluations were carried out using the default train/test splits from the UEA archive and hence allows us to easily compare with previously published benchmarks. The feature selection was carried out on the training set and the feature subset that was identified as the `best' subset was then used to evaluate on the test data. The computational cost was also recorded which is the time (sum of system and user CPU time) required to identify the `best' feature subset. The benchmark accuracy for the 1-NN DTW evaluations were taken from previously published results \cite{Bagnall2017} whilst the MiniRocket benchmarks were evaluated using the train/test data as published benchmark results are not yet available according to the authors knowledge. 

%\begin{figure}[h]
%\begin{subfigure}{.5\textwidth}
%  \centering
%  % include first image
%  \includegraphics[trim = 10 0 20 10, clip, width=1\linewidth]{Images/MSTS_datared.png}  
%  \caption{}
%  \label{fig:sub-first}
%\end{subfigure}
%\begin{subfigure}{.5\textwidth}
%  \centering
  % include second image
%  \includegraphics[trim = 10 0 20 10, clip,width=1\linewidth]{Images/GS_datared.png}  
%  \caption{}
%  \label{fig:sub-second}
%\end{subfigure}
%\caption{}
%\label{fig:datared}
%\end{figure}
\begin{figure}[h]
\begin{subfigure}{.5\textwidth}
  \centering
  % include first image
  \includegraphics[width=1.1\linewidth]{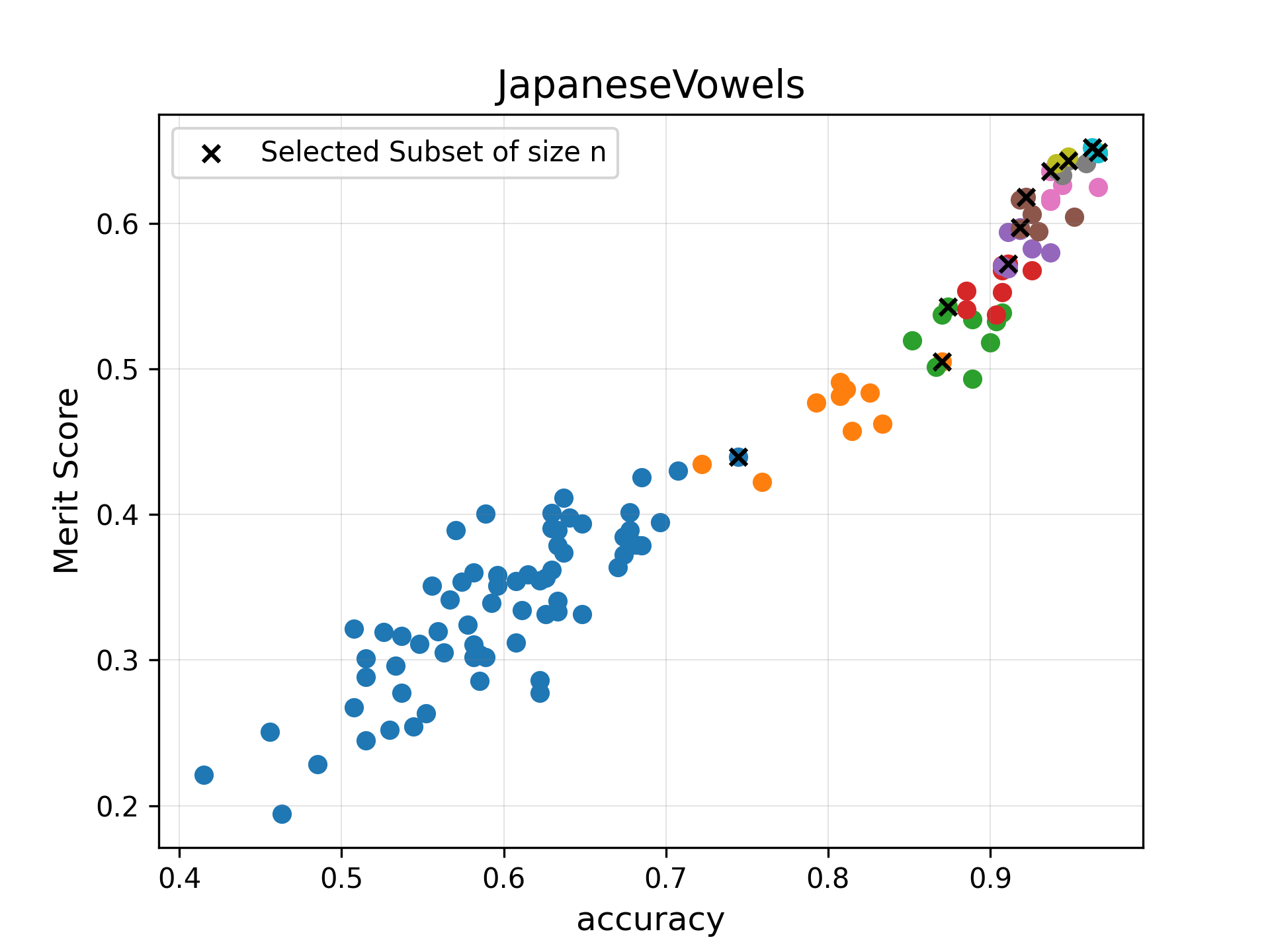}  
  \caption{JapaneseVowels - 12 dimensions [JV]}
  %\label{fig:sub-first}
\end{subfigure}
\begin{subfigure}{.5\textwidth}
  \centering
  % include second image
  \includegraphics[width=1.1\linewidth]{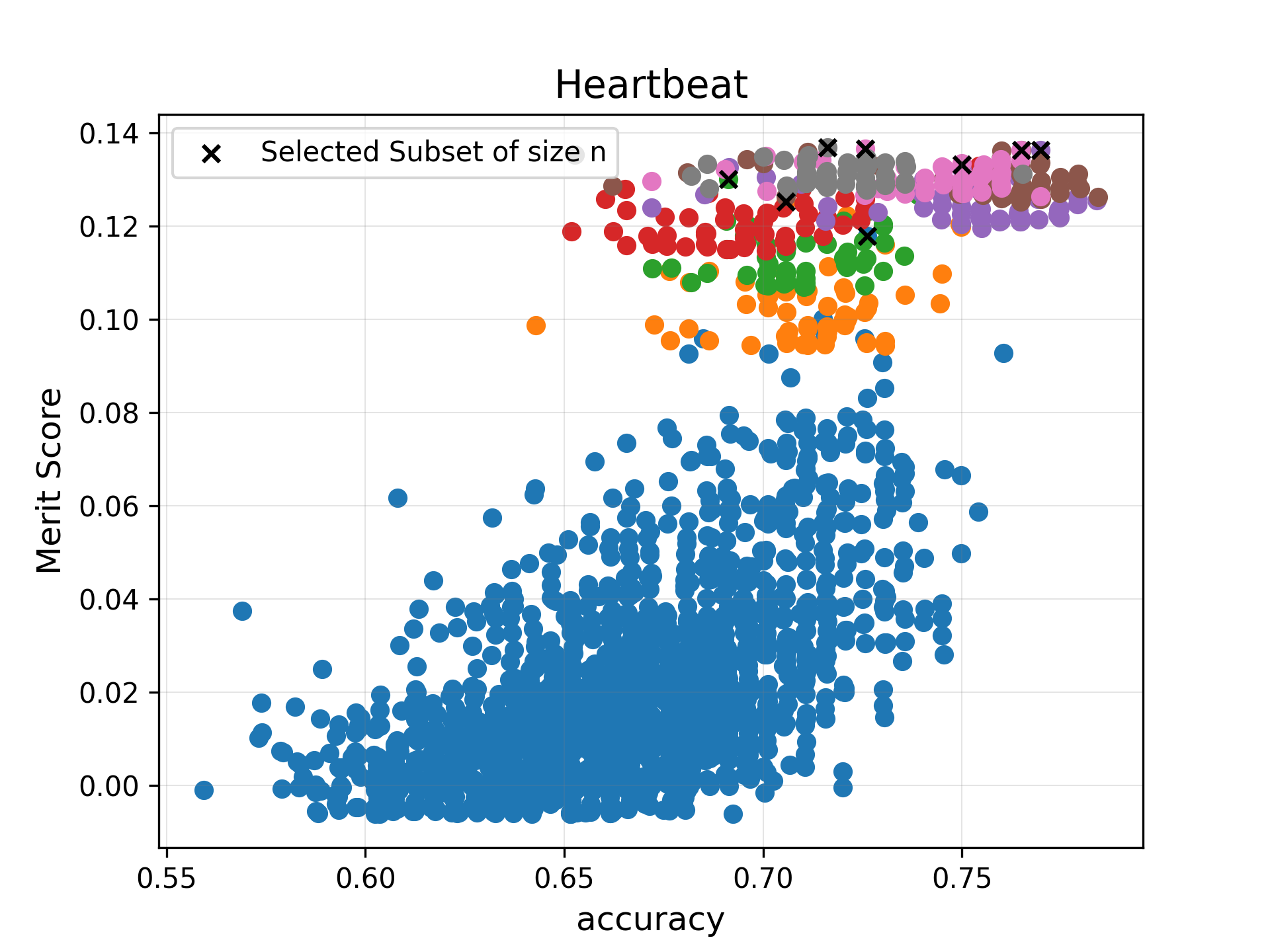}  
  \caption{Heartbeat - 61 dimensions [HB]}
  \label{fig:sub-second}
\end{subfigure}
\newline
\begin{subfigure}{.5\textwidth}
  \centering
  % include third image
  \includegraphics[width=1.1\linewidth]{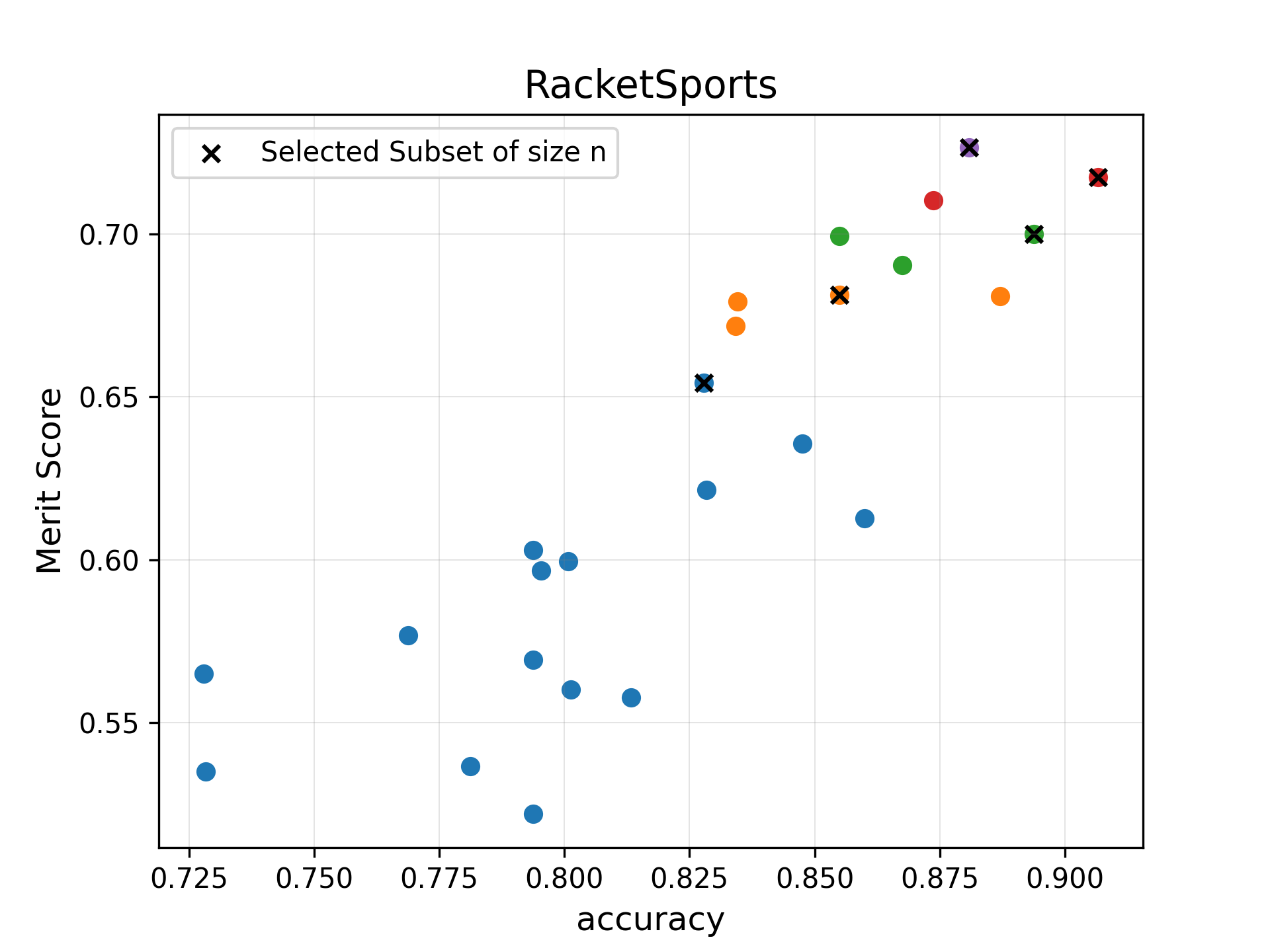}  
  \caption{RacketSports - 6 dimensions [RS]}
  %\label{fig:sub-third}
\end{subfigure}
\begin{subfigure}{.5\textwidth}
  \centering
  % include fourth image
  \includegraphics[width=1.1\linewidth]{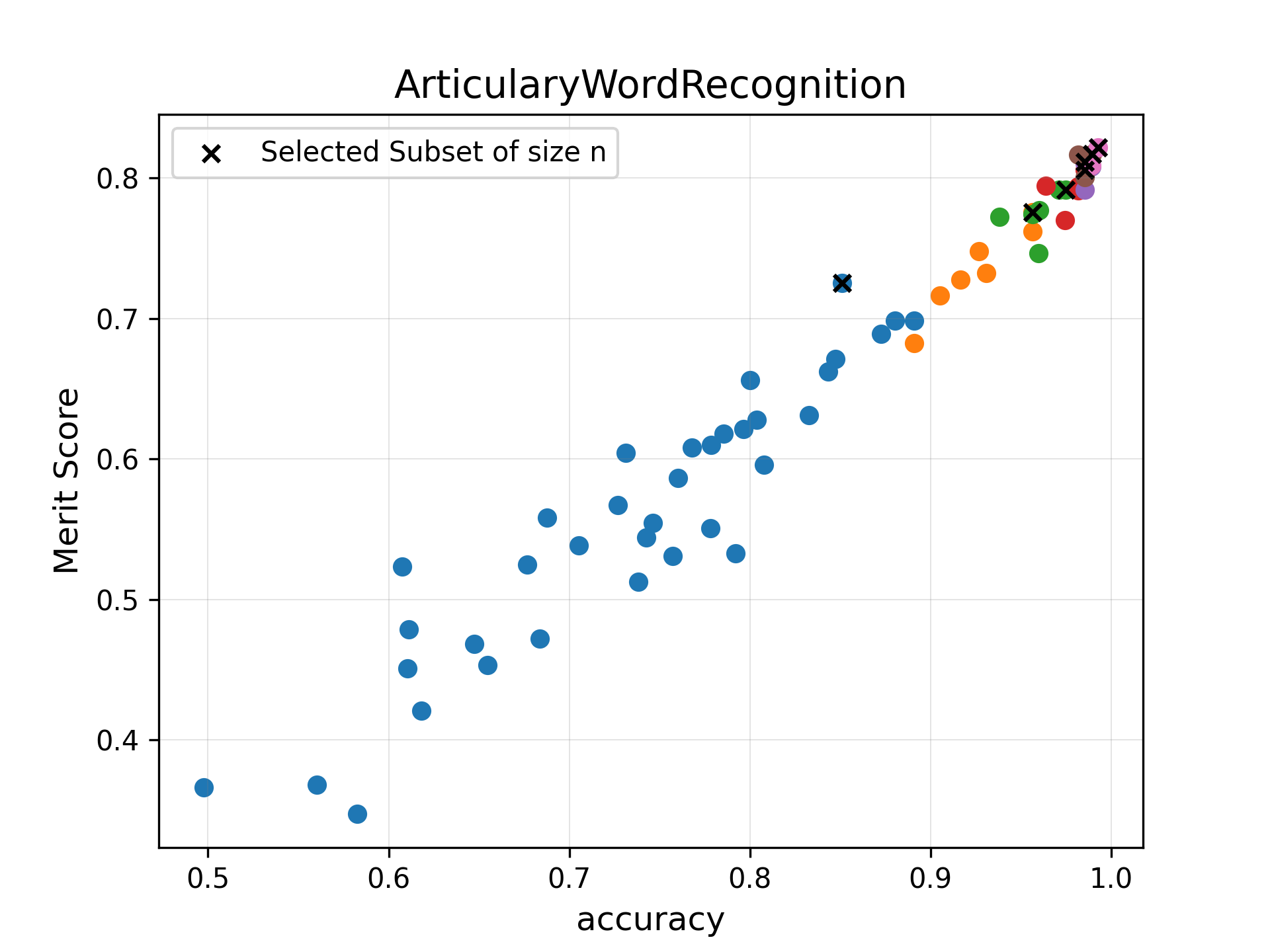}  
  \caption{ArticularyWordRecognition - 9 dimensions [AWR]}
  %\label{fig:sub-fourth}
\end{subfigure}
\caption{MSTS on a subset of UEA datasets, where the colours indicate different subset sizes and the cross indicates the subset selected for each subset size}
\label{fig:MSCOMP}
\end{figure}

\par
To obtain some insight into how well the Merit Score works in improving accuracy, we evaluate the accuracy from all the subsets considered by the MSTS approach for selected datasets. Figure \ref{fig:MSCOMP} shows the Merit Score obtained for four of the UEA datasets against the accuracy from the same feature subset when evaluated using a 1-NN DTW with 10 fold cross validation. This shows the Merit Score as evaluated using the training data and hence the feature subset with the highest Merit Score for each feature size would be the subset selected for at each step of the greedy search. The plots show the positive trend where as the Merit Score increases, the accuracy of that subset also increases. In some datasets, MSTS may be able to identify the importance of features for a classification task that are left behind in a Wrapper strategy. The HB dataset (Figure \ref{fig:sub-second}) shows how there may be merit to features even where it leads to a decrease in accuracy whilst training. The Wrapper strategy for the HB dataset had stopped at a 3-feature subset whereas the MSTS continued to find a 9-feature subset whilst maintaining an overall lower computational cost. For this HB dataset, the MSTS technique outperformed both the benchmark and the Wrapper strategy. This suggests that innate characteristics of certain datasets may make them more suitable for use with the MSTS technique. 
\section{Results \& Discussion}\label{Sec:Res}
The results from the evaluations are presented and discussed in this section. The effectiveness of the MSTS technique is evaluated in terms of classification accuracy, feature reduction, and processing time. The results are presented separately for the 1NN-DTW and MiniRocket classifiers and are detailed in the following sections. 
\subsection{1-NN DTW Evaluation}
A summary of the results obtained for the evaluations done using a 1NN-DTW classifier are given in Table \ref{tab:ResultsTableDTW}. The best or equal performing method in terms of classification accuracy out of MSTS, Wrapper, and the benchmark is given in bold. The feature selection techniques were used on the training data to find the `best' feature subset and this subset was used on the test data to get the classification accuracy. For the 1NN-DTW evaluations, three of the datasets (MI, Ph, SAD) were evaluated using just 60\% of the training data, as these had a large amount of training instances which would require a large amount of compute resources. To justify the use of 60\% of the data, smaller datasets were evaluated by gradually decreasing the amount of data used and analysing the effect of this on the overall accuracy. Using 60 \% of the data reduced the performance of the smaller datasets by less than 4\% and hence we hypothesise that the effect this will have on larger datasets would be smaller. As we use the same data for evaluating both MSTS and Wrapper approaches, taking a reduced subset of the training data should not bias the results and was done purely to reduce compute resources required as the alternative would be to remove these datasets from the evaluation. 
\par
Overall, the results suggest that when using a DTW classifier, MSTS and the Wrapper strategy perform nearly on par in terms of accuracy. MSTS outperforms Wrapper in 7 out of the 19 datasets and the Wrapper outperforms MSTS in 7 out of 19 datasets with both techniques performing on par for the remaining 5 datasets. In 11 out of the 19 datasets, MSTS is able to equal or improve the accuracy. In many of these cases, this is done with just a small subset of the original features. These results show that although some datasets may benefit more from the MSTS feature selection whilst others from the Wrapper strategy, only 5 of the 19 datasets have a higher accuracy from using all the features, hence suggesting that some level of feature selection would be beneficial whilst working with time series.
\par
The 1-NN DTW classifier used for the evaluation uses a distance matrix pre-computed for each dataset using dependent warping ($DTW_D$). The distance matrix for each feature was pre-computed and saved. This DTW matrix was saved in a 3D representation where each feature had a 2D matrix associated with it. The distance matrices for each feature is required for both MSTS and the Wrapper strategies and to calculate the distance matrix separately each time a new subset is evaluated would make the evaluation task nearly impossible in terms of the computational cost involved as even to evaluate the full distance matrix one time required weeks to compute for some of the larger datasets. Hence each time a subset is evaluated, the DTW distance would be calculated from looking-up the required distances from the pre-computed distance matrices. Hence the  times  reported here do not include the initial DTW distance compute time as this is common to both techniques which is an important point to note while interpreting the results, in particular when comparing with the MiniRocket results for which we do not store anything in advance. 
\par
When looking at computational expense, MSTS is faster for 16 datasets with the Wrapper only outperforming MSTS in 2 datasets. However noting that the DTW distance matrix was saved in advance, the true time required for the Wrapper could be higher. Hence MSTS clearly outperforms a Wrapper in terms of computational cost. 
\par
\begin{figure}
\centering
  \includegraphics[keepaspectratio, trim = 0 0 0 0, clip,width=0.8\textwidth]{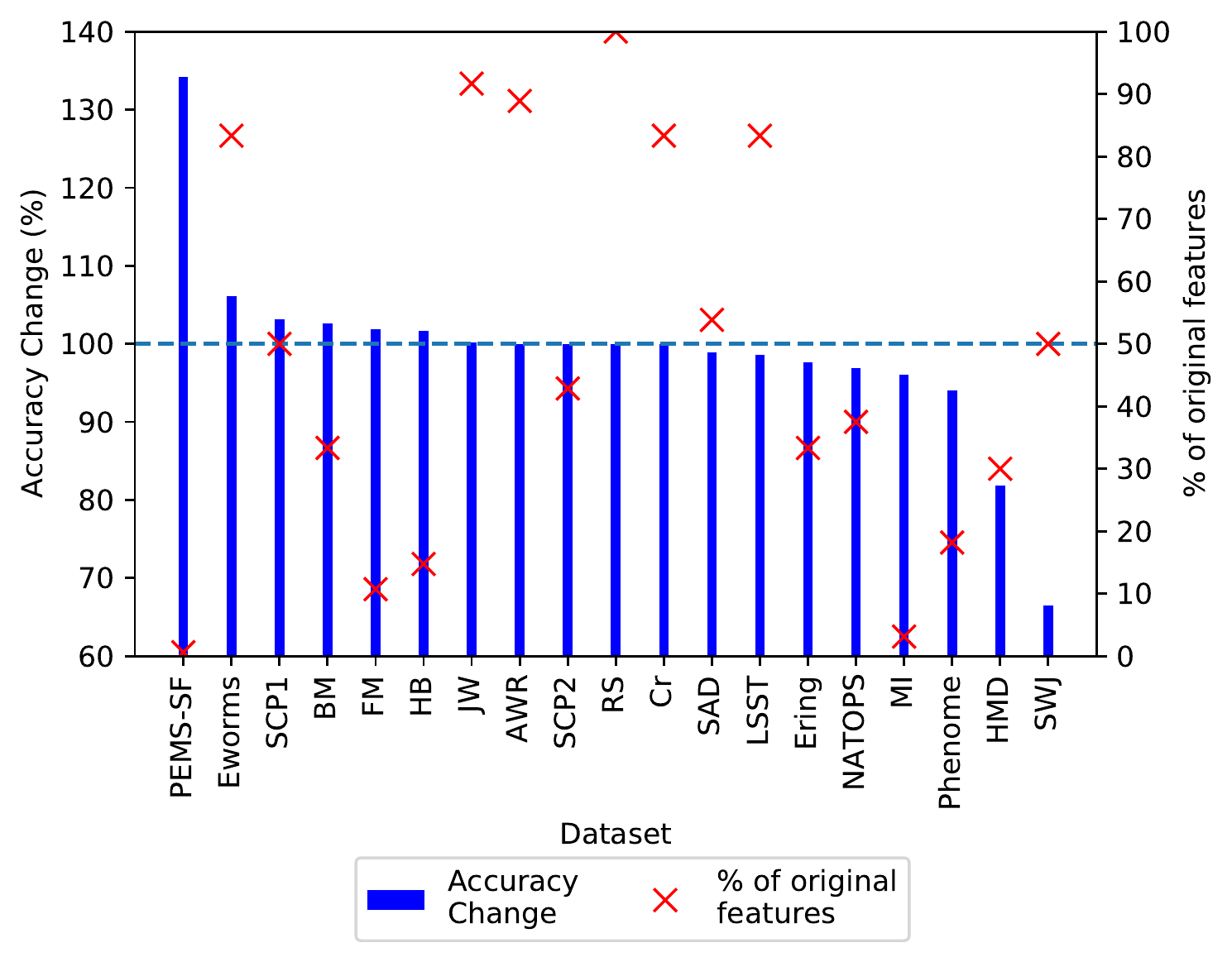}
\caption{Effect of MSTS on feature reduction and accuracy using a DTW classifier}
\label{fig:fullResDTW}
\end{figure}

Finally, in terms of feature reduction, both MSTS and Wrapper strategies are able to identify a reduced subset of features that are important for the task, with many datasets required less than half of the total features. Figure \ref{fig:fullResDTW} shows the accuracy change and the percentage of the original features that are selected using the MSTS feature selection technique. Here, it can be seen that PEMS-SF, Eworms, SCP1, BM, FM, and HB have an ideal case where the accuracy has increased while using a subset of the original features. There is then another 5 datasets with no change in accuracy, while using a subset of the original features (except RS which uses all 6 features). With the exception of the last two datasets; HMD and SWJ; each dataset only has a very small reduction in accuracy, if any at all while the reduction in the features required is quite significant for many of the datasets. It is also important to note that HMD and SWJ which are the two datasets where the MSTS causes a large reduction in the performance, had a very poor benchmark accuracy and hence the features may not be very good predictors and may not have much correlation between them. 
\begin{table}
%\begin{center}
\begin{tabular}{ x{1.45cm}|x{1.0cm}x{1.0cm}x{1.0cm}x{1.0cm}x{1.2cm}x{1.0cm}x{1.0cm}} 
\hline
 Dataset & Acc. (MSTS) & Acc. (Wr.) & Time Taken (MSTS) & Time Taken (Wr.) & Benchmark Acc. \cite{Bagnall2018} & Feat. Sel (MSTS) & Feat. Sel (Wr.) \\ 
\hline
AWR & \textbf{0.987} & \textbf{0.987} & 2.0967 & 2.122 & \textbf{0.987} & 8/9 & 8/9\\ 

FM & 0.54 & \textbf{0.57} & 1.991 & 12.675 & 0.53 & 3/28 & 2/28\\ 

JV & \textbf{0.951} & 0.9297 & 2.534 & 2.91 & 0.949 & 11/12 & 7/12\\ 

NATOPS & 0.856 & \textbf{0.889} & 6.599 & 7.606 & 0.883 & 9/24 & 5/24\\ 

RS & 0.803 & \textbf{0.842} & 0.313 & 0.551 & 0.803 & 6/6 & 3/6\\ 

SCP1 & \textbf{0.799} & 0.782 & 0.249 & 0.579 & 0.775 & 3/6 & 2/6\\ 

HB & \textbf{0.729} & 0.673 & 20.776 & 44.73 & 0.717 & 9/61 & 3/61\\ 

BM & \textbf{1} & \textbf{1} & 0.154 & 0.318 & 0.975 & 2/6 & 2/6\\

HMD & 0.189 & 0.149 & 0.48 & 1.249 & \textbf{0.231} & 3/10 & 2/10\\ 

Cr & \textbf{1} & \textbf{1} & 0.306 & 0.227 & \textbf{1} & 5/6 & 3/6\\

SWJ & 0.133 & \textbf{0.267} & 0.0846 & 0.0613 & 0.2 & 2/4 & 2/4\\ 

SCP2 & \textbf{0.539} & 0.533 & 0.266 & 0.637 & \textbf{0.539} & 3/7 & 2/7\\ 

PEMS-SF & \textbf{0.954} & 0.786 & 997.59 & 10472 & 0.711 & 6/963 & 3/693\\ 

LSST & 0.543 & 0.543 & 4.243 & 20.352 & \textbf{0.551} & 5/6 & 5/6 \\ 

EW & 0.656 & \textbf{0.679} & 0.2412 & 0.3927 & 0.618 & 5/6 & 2/6\\ 

ER & 0.893 & \textbf{0.915} & 0.047 & 0.047 & \textbf{0.915} & 2/6 & 4/6\\ 

MI & 0.48 & 0.46 & 4.66 & 23.05 & \textbf{0.5} & 2/64 & 2/64\\

Ph & 0.142 & 0.142 & 3.922 & 17.313 & \textbf{0.151} & 2/11 & 2/11\\
SAD & 0.952 & 0.956 & 16.24 & 155.43 & \textbf{0.963} & 7/13 & 9/13\\
\hline
\end{tabular}
\caption{Results from evaluation of MSTS and Wrapper strategies on UEA datasets with the best accuracy (or tied) in bold for the DTW classifier}
\label{tab:ResultsTableDTW}
%\end{center}
\end{table}
\subsection{MiniRocket Evaluation}
The MiniRocket evaluations were carried out using the default 10,000 kernels. A random state of 1995 was used for all experiments including the benchmark evaluations. A ridge regression classifier is used with the MiniRocket transformed data. The benchmark accuracy was evaluated using the default train/test split from the UEA archive as to the authors knowledge, these benchmark results for MiniRocket have not yet been published in the literature. The \textsf{sktime} implementation of MiniRocket for multivariate time series was used for this evaluation. 
\par
\begin{figure}
\centering
  \includegraphics[keepaspectratio, trim = 0 0 0 0, clip,width=0.8\textwidth]{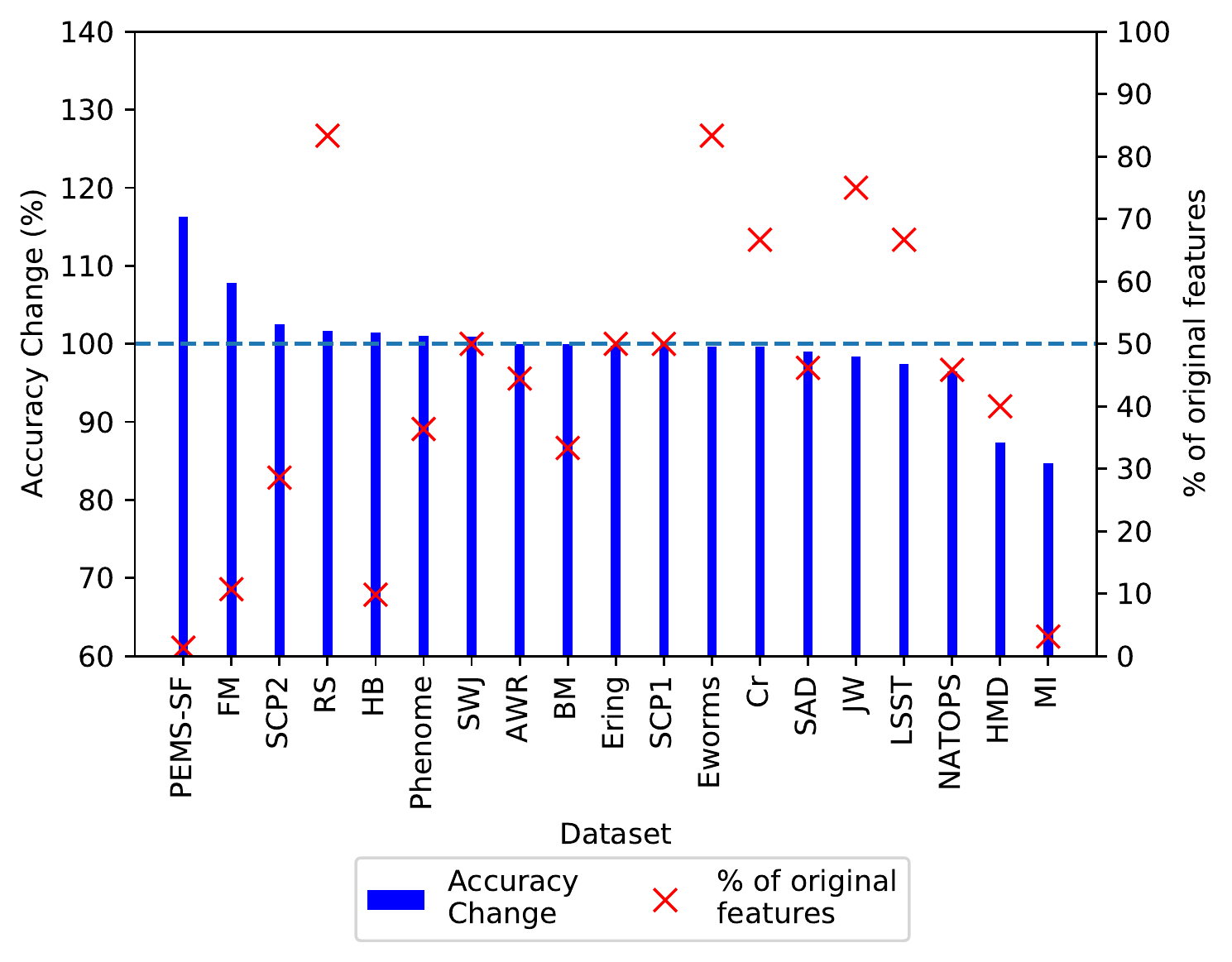}
\caption{Effect of MSTS on feature reduction and accuracy using a MiniRocket classifier}
\label{fig:fullResMiniRocket}
\end{figure}

The results for the MiniRocket evaluations are summarised in Figure \ref{fig:fullResMiniRocket}. Although the improvement in accuracy due to feature selection across the datasets is small when using the MiniRocket classifier, the accuracy is preserved for the majority of datasets with a significant reduction in the number of features required to obtain this accuracy.
As the MiniRocket classifier works well in identifying the more important features using its in-built functionality, the improvement in accuracy from doing feature selection if any is small.
\par
In improving the performance, the Wrapper strategy works well for the MiniRocket evaluations. As can be seen in Table \ref{ResultsTableMiniRocket}, 11 of the datasets are able to get best (or tied) performance using a Wrapper compared to MSTS and the benchmark, whereas MSTS is able to get the best (or tied) performance for 6 datasets. However, there is a considerable difference in the time taken with the MSTS technique performing better for every dataset except for the SWJ dataset which is however a small dataset (4 features). For larger datasets, MSTS presents a significant speedup with it being almost 115 times faster for MI and almost 1636 times faster for PEMS-SF which has 963 features. MSTS is particularly useful to improve the speed in larger datasets with more features such as MI (64 features), HB (61 features), and PEMS-SF (963 features) which have a more than 100 times speedup from using MSTS. From a feature reduction perspective, both MSTS and the Wrapper strategies work to similar extents with a similar number of features being selected. 
\begin{table}
%\begin{center}
\begin{tabular}{ x{1.45cm}|x{0.8cm}x{0.8cm}x{1.2cm}x{1.6cm}x{1.2cm}x{1.0cm}x{1.0cm}} 
\hline
 Dataset & Acc. (MSTS) & Acc. (Wr.) & Time Taken (MSTS) & Time Taken (Wr.) & Benchmark Acc. \cite{Bagnall2018} & Feat. Sel (MSTS) & Feat. Sel (Wr.) \\ 
\hline
AWR & \textbf{0.99} & \textbf{0.99} & 374 & 6565 & \textbf{0.99} & 4/9 & 4/9\\ 

FM &  0.55 & \textbf{0.63} & 1258 & 37826 & 0.51 & 3/28 & 2/28\\ 

JV &  0.974 & 0.974 & 478 & 8576 & \textbf{0.99} & 9/12 & 6/12\\ 

NATOPS & 0.878 & \textbf{0.917} & 803 & 23740 & 0.91 & 11/24 & 4/24\\ 

RS & \textbf{0.895} &  0.869 & 199 & 1542 & 0.88 & 5/6 & 3/6\\ 

SCP1 & 0.918 & \textbf{0.925} & 254 & 3004 & 0.92 & 3/6 & 2/6\\ 

HB & 0.761 & \textbf{0.785} & 2251 & 225515 & 0.75 & 6/61 & 3/61\\ 

BM & \textbf{1} & \textbf{1} & 17 & 128 & \textbf{1} & 2/6 & 2/6\\

HMD &  0.297 & \textbf{0.419} & 314 & 5325 & 0.34 & 4/10 & 2/10\\ 

Cr & 0.986& 0.972 & 169 & 1198 & \textbf{0.99} & 4/6 & 2/6\\

SWJ &  \textbf{0.333} & \textbf{0.333}  & 9.938  & 2.203 & 0.33 & 2/4 & 2/4\\ 

SCP2 & \textbf{0.533} & 0.528 & 251 & 3386 & 0.52 & 2/7 & 2/7\\ 

PEMS-SF & 0.965 & \textbf{0.988} & 32084 & 52503012 & 0.83 & 13/963 & 2/963\\ 

LSST & 0.633 & 0.641 &  2240 & 26216 & \textbf{0.65} & 4/6 & 4/6 \\ 

EW & 0.947 & 0.939 & 173 & 4035 & \textbf{0.95} & 5/6 & 3/6\\ 

ER & 0.978 & \textbf{0.985} & 11 & 34 & 0.98 & 2/4 & 2/4\\ 

MI & 0.5 & 0.5 & 2909 & 335972 &\textbf{0.59} & 2/64 & 3/64\\

Ph & \textbf{0.293} & \textbf{0.293} & 7070 & 149483 & 0.29 & 4/11 & 2/11\\
SAD & 0.99 & 0.99 & 36102 & 1143822 & \textbf{1} & 6/13 & 8/13\\
\hline
\end{tabular}
\caption{MiniRocket classifier Results of MSTS and Wrapper strategies on UEA datasets with the best accuracy (or tied) in bold}\label{ResultsTableMiniRocket}
%\end{center}
\end{table}
\subsection{CFS for Time Series}
The results suggest that CFS could be a good technique for feature subset selection of time series in particular for the purpose of data reduction. When working outside of the time series domain, CFS has been known to favour small feature subsets with moderate accuracy even where there is some merit to adding more features \cite{Cunningham2021}. However, this effect is not as evident whilst working with time series data. In our results, the CFS technique tends to select similar or larger feature subsets than the Wrapper approach. This suggests that CFS may be more suited for time series than tabular data perhaps due to greater correlations between features when working with time series data.
\par
It is worth comparing these results with the feature subset selection technique proposed by Ircio \textit{et al.,} which uses mutual information based on $k$-NN as it is directly comparable \cite{Ircio2020}. 
The authors show that on average for 8 out of the 16 datasets for $DTW_D$ and 4 out of the 16 datasets for $DTW_I$ the accuracy can be maintained or improved \cite{Ircio2020}. In our evaluations, MSTS improves or maintains the accuracy for 11 out of the 19 datasets for DTW and for 12 out of the 19 datasets for MiniRocket. Furthermore, it is observed that it is for some of our worst performing datasets such as MI and HMD that the $k$-NN based method works best. Similarly, our best performing dataset, PEMS-SF has a poorer performance using the $k$-NN approach. This suggests that MSTS may be better in maintaining good performance where the initial predictive capabilities of the dataset are good whereas other techniques such as the $k$-NN based method of identifying a feature subset may be preferred for datasets where the initial predictive capabilities are not great.
\par
The evaluations also show that the MiniRocket classifier is clearly dominant over 1-NN-DTW and as a result the MiniRocket Wrapper evaluations give a better accuracy improvement than the DTW Wrapper strategy. Although Wrapper works well with a good classifier, MSTS is less dependent on how well the classifier works for the task. Overall, for both classifiers, MSTS maintains the accuracy for the majority of datasets and of those where the accuracy is not preserved such as MI and HMD, they generally have low overall accuracy even with the full set of data suggesting low predictive capability in the data.
\section{Conclusions}
\label{sec:conc}
The evaluations carried out suggest that the proposed MSTS technique works well for feature subset selection. The proposed MSTS technique particularly shows potential as a feature reduction technique as it is able to dramatically reduce the number of features required whilst maintaining the accuracy and is able to do this with less computational expense than what is required for a Wrapper strategy.
\par
Nearest neighbour based techniques are known to be more affected by noise \cite{Schafer2016} and hence may benefit more from data reduction. This is evident in our results as the improvement in accuracy when reducing features prior to using a 1NN-DTW classifier is greater than that from using the MiniRocket classifier. Furthermore, Rocket based classifiers have a level of feature extraction in-built in them meaning any feature selection on top of this may not be as beneficial in improving accuracy. However, regardless of the classifier being used, feature subset selection serves multiple purposes and the results have shown the MSTS is valuable as a feature reduction technique as in many cases, the feature subset size can be reduced dramatically whilst maintaining or improving accuracy.
\par
The choice of technique to use would be dependent on the application and data. Where feature reduction is not necessary, MiniRocket alone with its feature selection capabilities may provide a good performing model. However, where a subset of features is sought either to reduce future data collection or for explainability purposes, MSTS is recommended for use. MSTS is cost effective compared to the Wrapper and is able to select the important features as well as the Wrapper technique. Hence the MSTS technique is valuable for feature reduction.

\section*{Acknowledgements}
This work has emanated from research conducted with the financial support of Science Foundation Ireland under the Grant number 18/CRT/6183. For the purpose of Open Access, the author has applied a CC BY public copyright license to any Author Accepted Manuscript version arising from this submission.

\bibliographystyle{plain}  
  % APS-like style for physics
\bibliography{FSS}
\end{document}